\documentclass[10pt,conference]{IEEEtran}
\usepackage{amsmath}  

\usepackage[utf8]{inputenc}
\usepackage[switch]{lineno}
\usepackage{floatrow}
\usepackage{amsfonts}
\usepackage{graphicx}
\usepackage{longtable}
\usepackage{caption}
\usepackage{subcaption}
\usepackage{placeins}
\usepackage[tableposition=top]{caption}

\usepackage{floatrow}
\usepackage{changes}
\definechangesauthor[name={parsa karbasi}, color=red]{PK}

\usepackage{listings, xcolor}

\definecolor{verylightgray}{rgb}{.97,.97,.97}

\lstdefinelanguage{Solidity}{
	keywords=[1]{anonymous, assembly, assert, balance, break, call, callcode, case, catch, class, constant, continue, constructor, contract, debugger, default, delegatecall, delete, do, else, emit, event, experimental, export, external, false, finally, for, function, gas, if, implements, import, in, indexed, instanceof, interface, internal, is, length, library, log0, log1, log2, log3, log4, memory, modifier, new, payable, pragma, private, protected, public, pure, push, require, return, returns, revert, selfdestruct, send, solidity, storage, struct, suicide, super, switch, then, this, throw, transfer, true, try, typeof, using, value, view, while, with, addmod, ecrecover, keccak256, mulmod, ripemd160, sha256, sha3}, 
	keywordstyle=[1]\lst@ifdisplaystyle\color{blue}\bfseries \fi,
	keywords=[2]{address, bool, byte, bytes, bytes1, bytes2, bytes3, bytes4, bytes5, bytes6, bytes7, bytes8, bytes9, bytes10, bytes11, bytes12, bytes13, bytes14, bytes15, bytes16, bytes17, bytes18, bytes19, bytes20, bytes21, bytes22, bytes23, bytes24, bytes25, bytes26, bytes27, bytes28, bytes29, bytes30, bytes31, bytes32, enum, int, int8, int16, int24, int32, int40, int48, int56, int64, int72, int80, int88, int96, int104, int112, int120, int128, int136, int144, int152, int160, int168, int176, int184, int192, int200, int208, int216, int224, int232, int240, int248, int256, mapping, string, uint, uint8, uint16, uint24, uint32, uint40, uint48, uint56, uint64, uint72, uint80, uint88, uint96, uint104, uint112, uint120, uint128, uint136, uint144, uint152, uint160, uint168, uint176, uint184, uint192, uint200, uint208, uint216, uint224, uint232, uint240, uint248, uint256, var, void, ether, finney, szabo, wei, days, hours, minutes, seconds, weeks, years},	
	keywordstyle=[2]\lst@ifdisplaystyle\color{teal}\bfseries \fi,
	keywords=[3]{block, blockhash, coinbase, difficulty, gaslimit, number, timestamp, msg, data, gas, sender, sig, value, now, tx, gasprice, origin},	
	keywordstyle=[3]\lst@ifdisplaystyle\color{violet}\bfseries \fi,
	identifierstyle=\color{black},
	sensitive=false,
	comment=[l]{//},
	morecomment=[s]{/*}{*/},
	commentstyle=\color{gray}\ttfamily,
	stringstyle=\color{red}\ttfamily,
	morestring=[b]',
	morestring=[b]"
}

\usepackage[bookmarks=false]{hyperref}
\modulolinenumbers[5]
\usepackage{subcaption}
\usepackage{graphicx}
\usepackage{multirow}
\usepackage{pifont}

\begin{document}


\title{Effective Black Box Testing of Sentiment Analysis Classification Networks}



\author{\IEEEauthorblockN{Parsa Karbasizadeh\IEEEauthorrefmark{1},
		Fathiyeh Faghih\IEEEauthorrefmark{2}, and Pouria Golshanrad\IEEEauthorrefmark{3} \\}
	\IEEEauthorblockA{College of Engineering, Department of Electrical and Computer Engineering \\ University of Tehran, Tehran, Iran \\
		Email: \IEEEauthorrefmark{1}parsa.karbasi@ut.ac.ir,
		\IEEEauthorrefmark{2}f.faghih@ut.ac.ir,
		\IEEEauthorrefmark{3}pouria.golshanrad@ut.ac.ir}}

	\maketitle
	
%
\begin{abstract}
Transformer-based neural networks have demonstrated remarkable performance in natural language processing tasks such as sentiment analysis. Nevertheless, the issue of ensuring the dependability of these complicated architectures through comprehensive testing is still open. This paper presents a collection of coverage criteria specifically designed to assess test suites created for transformer-based sentiment analysis networks. Our approach utilizes input space partitioning, a black-box method, by considering emotionally relevant linguistic features such as verbs, adjectives, adverbs, and nouns. In order to effectively produce test cases that encompass a wide range of emotional elements, we utilize the k-projection coverage metric. This metric minimizes the complexity of the problem by examining subsets of k features at the same time, hence reducing dimensionality. Large language models are employed to generate sentences that display specific combinations of emotional features. The findings from experiments obtained from a sentiment analysis dataset illustrate that our criteria and generated tests have led to an average increase of 16\% in test coverage. In addition, there is a corresponding average decrease of 6.5\% in model accuracy, showing the ability to identify vulnerabilities. Our work provides a foundation for improving the dependability of transformer-based sentiment analysis systems through comprehensive test evaluation. 
\end{abstract}

\markboth{Journal of IEEE Software}%
{Shell \MakeLowercase{\textit{et al.}}: Testing Smart Contracts Gets Smarter}





\section{Introduction}\label{sec:introduction}

The impact of neural networks on various fields, including image recognition, natural language processing, and self-driving cars, has been significant~\cite{wu}. In the context of Natural Language Processing (NLP), neural networks have gained notable recognition due to their ability to enhance accuracy~\cite{Chen_2020}.
Sentiment analysis, also known as opinion mining or emotion AI, is a natural language processing (NLP) technique used to determine the emotional tone or subjective opinion expressed in text.  This procedure utilizes NLP methodologies to analyze and interpret human language, enabling diverse applications across various domains. For instance, in marketing, sentiment analysis can be leveraged to evaluate customer opinions about products and services, providing valuable insights for brand reputation management. 
Various tools, including NLTK~\cite{bird2006nltk}, are employed to interpret customer sentiment and capture embedded emotions. 

Traditional emotion recognition methods often struggle with the inherent complexities of human language. These complexities include word ambiguity, where a word's emotional connotation can shift depending on context (e.g., ``scared'' can be positive in an exciting situation or negative in a threatening one). In addition, sarcasm and negation pose challenges in identifying true sentiment.  Furthermore, the process of understanding emotional meaning must include evaluating words that are not directly close to each other in a sentence, which is a step away from traditional methods.

Transformer-based neural networks effectively handle the ambiguity in emotion sentiment analysis through their attention mechanisms and large parameters~\cite{vaswani2017attention}. Parallel text processing allows them to analyze text data efficiently and capture the hidden emotional factors that can exist across longer distances within a sentence. Through pre-training on extensive datasets, transformers are capable of learning complex language patterns and the subtle details of human emotional expression~\cite{vaswani2017attention}.

To guarantee the efficacy of these complex models, comprehensive testing protocols are essential. A significant challenge lies in evaluating the sufficiency of test suites specifically designed for Transformer-based emotional sentiment analysis networks. Given their complex architecture and the large number of parameters, conventional methods like neuron coverage~\cite{deepgauge} seem to be impractical. To address this limitation, we introduce in this work a novel set of coverage criteria designed to evaluate test suites for the assessment of Transformer-based emotion sentiment analysis networks. We employ the principle of input space partitioning, segmenting textual data into emotionally-relevant categories. This strategy enables us to concentrate on particular areas of the input space and generate targeted test cases.

To achieve high coverage with the least number of test cases, we employ the k-projection coverage method~\cite{k-project}. This technique utilizes dimensionality reduction by examining smaller feature sets at the same time. Concentrating on these smaller subsets allows for an efficient evaluation of the network's performance across diverse emotional categories within the input space. This strategy ensures comprehensive coverage with fewer tests, thereby eliminating the necessity for extensive test suites.
Additionally, we introduce a set of specialized features specifically designed for emotional sentiment analysis. These features capture the emotional details within text data, thereby aiding the network in accurately identifying the intended emotions.

Upon applying our emotion-specific k-projection coverage method to a dataset designed for emotional sentiment analysis, we observed a 16\% average increase in coverage of the generated sentences. This enhanced coverage led to 6.5\% decrease in the accuracy of the System-Under-Test (SUT) on the new test suites, demonstrating the effectiveness of our approach in uncovering additional faults.

\textbf{Related Work:}This paper explores concepts related to test generation, black-box testing, and model coverage in DNN models. It summarizes the current state of coverage and performance measures for DNN models and discusses previous research on utilizing diversity in black-box DNN testing to evaluate DNNs and conventional software systems \cite{10,deepgauge}.

Transformers are preferred over traditional DNNs due to their superior accuracy, particularly in AI domains like sentiment analysis \cite{bhadresh_savani_distilbert_base_uncased_emotion,jorgeutd2023sagemakerrobertabaseemotion}. Evaluating the adequacy of test suites for emotional sentiment analysis networks poses a significant challenge. In software engineering, test suite effectiveness is assessed through black-box and white-box testing \cite{henard2016comparing}. Applying white-box criteria to transformer-based NLP systems is limited by their complexity and internal access requirements. Alternative testing methods such as black-box strategies or input diversity metrics may be necessary to ensure their reliability \cite{zohre}.

Input space partitioning is an effective approach for black-box testing, particularly in text data, where emotionally relevant features can be used to partition the input domain into smaller subsets for precise test case design \cite{Myers2016}. 

$k$-projection coverage, employing dimensionality reduction, shows promise in image recognition for evaluating self-driving car perception systems using CNN-based neural networks and GAN-based models \cite{k-project}. However, its potential in text-based emotional sentiment analysis remains unexplored.

The contributions of our study are as follows: 
\begin{enumerate}
    \item We introduce novel coverage criteria based on $k$-projection for black-box testing of sentiment analysis systems. This metric serves a dual purpose: evaluating the effectiveness of existing test suites and guiding the generation of new test cases to augment them.
\item Our experimental results demonstrate that enriching test suites using this proposed coverage criterion leads to a significant reduction in model accuracy. This finding has two key implications:
\begin{itemize}
    \item The coverage criterion is a valid metric for assessing test suites in this context, as higher coverage scores correlate with a greater ability to uncover faults and decrease model accuracy.
\item The coverage criterion can be effectively utilized to generate effective test cases that reveal failures and thus improve the overall quality of test suites.
\end{itemize}

\end{enumerate}

The remainder of this paper is organized as follows. We explain the proposed methodology in Section~\ref{sec:methodology}, followed by a presentation of the experimental results in Section~\ref{ssec:evaluation}.  Section~\ref{sec:conclusion} concludes the paper by summarizing our findings and future work.

\section{Methodology}\label{sec:methodology}

The performance of a neural network is typically evaluated by calculating its accuracy on a test suite. However, accuracy alone is not a sufficient measure of model quality. The effectiveness of testing and the reliability of the results heavily depend on the adequacy of the test suite. Coverage criteria are metrics used to assess how effectively a test suite exercises different aspects of a system under test. In traditional software testing, common criteria include statement, branch, and path coverage~\cite{branchcover}. However, these traditional criteria cannot be directly applied to learning models due to their fundamentally different architecture~\cite{zohre}. To overcome this challenge, we propose a set of black-box metrics for evaluating test suites specifically designed for sentiment analysis tasks.

Our approach originates from Input Space Partitioning (ISP), a black-box software testing technique. ISP involves dividing a system's input domain into distinct subsets or partitions, where each partition represents inputs expected to result in similar system behavior. Test cases are then strategically selected from each partition to ensure comprehensive coverage of the input space.

\subsection{Input Space Partitioning}
\label{sec:ISP}

The primary objective of this study is to assess the effectiveness of utilizing the input space partitioning technique to evaluate test suites for emotional sentiment analysis in Transformer-based neural networks. This technique utilizes the features of the input data to partition the input space into distinct regions, facilitating organization and understanding of the data. \cite{k-project}

The architecture of the neural network model and the nature of the input data require the identification and selection of appropriate features for input space partitioning. Effective evaluation of a test suite requires developing a feature space that covers a wide range of scenarios and real-world instances. 
An effective feature space can be established by adherence to the guidelines proposed in Cheng et al.'s study \cite{k-project}, which outlines specific autonomous driving scenarios that the test cases should encompass and which will be defined subsequently. 

To establish a clear foundation for the framework discussed in this study, we introduce the following notations. These notations are essential to understanding the subsequent mathematical definitions and analyses.
\begin{enumerate}

\item \textbf{Input Domain}:
The input domain includes all textual inputs the neural network can receive. All text variations the network must process and respond to are included.
    \[
    D = \{d_1, d_2, \ldots, d_n\}
    \]
where \(d_i\) represents an individual input text.
\item \textbf{Partition}:
In order to understand the existing components and assess the test suite, it is crucial to establish specific partitions and map the input domain within those defined partitions. In this study, partitions are defined as follows:
\[
    P = \{P_1, P_2, \ldots, P_k\}
    \]
    where \(P_i \subseteq D\) and \(P_i \cap P_j = \emptyset\) for \(i \neq j\), and \(\bigcup_{i=1}^k P_i = D\).
    
\item \textbf{Features}:
Through the analysis of the input domain and the consideration of the functionality of the system under test (SUT), we can determine the important features that will be used to establish partitions.
\[
    F = \{F_1, F_2, \ldots, F_m\}
    \]
    
\item \textbf{Test Suite}: Representative tests should be selected from each partition to form a test suite.
    \[
    T = \{t_1, t_2, \ldots, t_l\}
    \]
    where \(t_i \in P_j\) for some \(P_j \in P\).
\end{enumerate}

\subsection{Emotional Features}
\label{subsec2.2.1}
Input space partitioning relies on specific features to establish partitions, crucial for evaluating test suite effectiveness in neural networks. The upcoming section will outline the specific features employed in this study.

The meaning and sentiment of a sentence can be identified through crucial components such as verbs, adverbs, adjectives, and other linguistic elements. Therefore, features are determined according to these important elements of a sentence.

Verbs strongly influence a sentence's sentiment. For instance, in "She laughed when I saw her," the verb "laughed" implies joy, highlighting its importance in sentiment analysis.

Besides the verb, the adverb in a sentence is another influential feature affecting its sentiment. For example, in "He bites his nails nervously," the adverb "nervously" alone can convey fear or anxiety, making its presence significant for sentence categorization.

Adjectives shape a sentence's emotional tone by describing specific feelings. For example, "She received a joyful gift" conveys happiness, whereas "She received a disappointing gift" indicates sadness.

Nouns contribute significantly to a sentence's emotion. In "The old man sat alone, his eyes filled with tears, mourning the loss of his dearest friend," nouns like "tears" and "loss" contribute to its "sad" emotional label.

In this research, the emotional attributes of nouns were integrated as an additional feature alongside verbs, adjectives, and adverbs.
The sets of features presented in this paper are denoted as:

\[
F = \{S_V, S_{ADJ}, S_{ADV}, S_N\}
\]

where:
\begin{itemize}
    \item \(S_V\) represents sentiment scores for verbs, categorized into emotional labels: \[
    S_V : V \rightarrow \{\text{joy}, \text{anger}, \text{sadness}, \text{fear}, \text{surprise}\}
    \]
    \item \(S_{ADJ}\) represents sentiment scores for adjectives,  categorized into emotional labels: \[
    S_{ADJ} : ADJ \rightarrow \{\text{joy}, \text{anger}, \text{sadness}, \text{fear}, \text{surprise}\}
    \]
    \item \(S_{ADV}\) represents sentiment scores for adverbs,  categorized into emotional labels: \[
    S_{ADV} : ADV \rightarrow \{\text{joy}, \text{anger}, \text{sadness}, \text{fear}, \text{surprise}\}
    \]
    \item \(S_N\) represents sentiment scores for nouns, categorized into emotional labels, including a neutral category: \[
    S_N : N \rightarrow \{\text{joy}, \text{anger}, \text{sadness}, \text{fear}, \text{surprise}, \text{neutral}\}
    \]
\end{itemize}

When dealing with multiple features and their numerous potential values, the number of possible combinations can grow exponentially. This can make generating comprehensive input data, encompassing all features at once, extremely complex. $k$-projection coverage is a technique used in input space partitioning to overcome this challenge. It aims to systematically explore the input space by focusing on a subset of `$k$' parameters at a time, while keeping the remaining parameters fixed.

To illustrate this method, consider an example of four features $\{F_1, F_2, F_3, F_4\}$, each taking two values $\{0, 1\}$. resulting in a total of $2^4 = 16$ possible combinations.
Applying $k$-projection coverage with different values of $k$ results in the following:
\begin{itemize}
    \item $k = 1$ (Single Feature Projection): We focus on one feature at a time, keeping the others fixed.
\begin{itemize}
    \item Number of projections: 4 (one for each feature)
    \item Number of cases per projection: 2 (the two possible values of the feature)
\end{itemize}
\item $k = 2$ (Pairwise Projection): We focus on pairs of features, keeping the other two fixed.
\begin{itemize}
    \item Number of projections: 6 (combinations of 4 features taken 2 at a time)
\item Number of cases per projection: 4 (2 values for each of the two features)
\end{itemize}
\item $k = 3$ (Three-way Projection): We focus on groups of three features, keeping the remaining one fixed.
\begin{itemize}
    \item Number of projections: 4 (combinations of 4 features taken 3 at a time)
\item Number of cases per projection: 8 (2 values for each of the three features)
\end{itemize}
\item $k = 4$ (All Features Projection): We consider all features simultaneously.
\begin{itemize}
    \item Number of projections: 1
\item Number of cases per projection: 16 (all possible combinations)
\end{itemize}
\end{itemize}

 Expanding on the previous example, this approach systematically selects sets of \emph{k} features and strives to achieve comprehensive coverage within a newly formed reduced dimension. This technique of dimensionality reduction enables efficient test generation, as it requires fewer test cases to achieve complete coverage of the input space.
 
 Generally, to calculate the coverage rate for $k$ = n, where n is the number of problem features, the following formula is employed:

\begin{equation}
\text{COV} = \frac{C(D)}{\alpha^n}
\end{equation}
In the above equation, \(C(D)\) represents the coverage rate of the test data over possible states, where \(\alpha\) is the range of each feature and \(n\) is the problem's dimensionality. In our context, \(\alpha=6\) and \(n=4\), requiring 1296 unique test data points for full coverage. This approach, however, is computationally demanding and requires substantial test data.

 Alternatively, when considering the effect of $k$, the following formula is used:

\begin{equation}
\label{cov_equation}
\text{COV} = \frac{C(D)}{\binom{n}{k}\alpha^k}
\end{equation}

 Considering the inverse correlation of the denominator, it is evident that the denominator, which is affected by variables such as \emph{k}, \(\alpha\), and \emph{n}, greatly decreases the number of tests needed to achieve complete coverage. 

 Incorporating all four features in sentences can be challenging and costly, reducing sentence clarity. Thus, this approach is ineffective. In order to tackle this issue, the dimensions of the feature coverage space are reduced, with the variable \(k\) being set to values of 2 and 3 in this study.The values for the variable $k$ are compared in Section \ref{ssec:evaluation} to evaluate their individual effects.

After measuring the coverage achieved by the test suite to assess the effectiveness of the $k$-projection metric, we attempt to increase coverage by generating test cases for uncovered scenarios. Ultimately, we explore whether improving coverage aids in the detection of hidden flaws in existing sentiment analysis networks.

\subsection{Evaluation of the Coverage Criteria}
Our approach to evaluating the proposed coverage criterion involves examining whether increasing a test suite's coverage score uncovers more failures in the sentiment analysis model. We introduce a method for augmenting test suites with additional test cases specifically designed to improve their coverage score. This evaluation method has two key advantages:
\begin{itemize}
    \item \textbf{Validation of Coverage Criteria}: Observing a decrease in model accuracy when using the augmented test suite confirms the effectiveness of the coverage criteria for evaluating sentiment analysis test suites. A higher coverage score indicates a greater ability to identify faults and lower model accuracy, thus validating the criteria.
    \item \textbf{Practical Test Case Generation}: The proposed method offers practitioners a valuable tool for generating effective test cases in sentiment analysis tasks. By focusing on increasing coverage, practitioners can systematically identify and address potential weaknesses in their models.
\end{itemize}

To uncover unexplored regions within the test suite based on our proposed coverage criteria, we propose a strategy for generating novel sentences. Manual sentence creation, while a viable method for producing desired test cases, is both time-intensive and unsuitable for generating large quantities.

Our idea in this study is to use Large Language Models (LLMs)~\cite{LMM} to generate the required sentences. LLMs require training using huge quantities of textual data, which allows them to imitate human-like text production. Examples of such LLMs include GPT-3~\cite{GPT} and LaMDA~\cite{lamda}. In this study, we employed Claude 3 Opus~\cite{anthropic2023}, an LLM produced by Anthropic, which is recognized for its exceptional sentence generation capabilities.

When utilizing LLMs for test generation, explicit prompts play a crucial role in guiding the model to produce specific types of test data that align with the research objectives. By evaluating the datasets using defined features in \ref{sec:methodology}, it is apparent that this dataset does not contain inputs with mixed emotions. For instance, after analyzing the test suites, we identified that the test suite lacks sentences where a verb is labeled with \emph{sadness} and an adverb is labeled with \emph{joy}. We generate such a test case using the following formulaic approach:

Let \( S \) represent the set of all sentences in the test suite, and define subsets \( V \) and \( A \) as follow:
\[
V = \{s \in S \mid \text{verb}(s) = \text{sadness}\}
\]
\[
A = \{s \in S \mid \text{adverb}(s) = \text{joy}\}
\]
The required test case would be $s' \in V \cap A$. To generate the missing test case \( s' \), we utilize:
\[
s' = \text{LLM}(\text{"Generate a sentence with a verb labeled as sadness} 
\]
\[
\quad \quad \quad \text{and an adverb labeled as joy"})
\]

This approach ensures that the generated sentence \( s' \) fills the identified gap in the test suite, thereby increasing the coverage of the test suite according to our proposed criteria. By adding \( s' \) to \( S \), we obtain an augmented test suite \( S' \):
\[
S' = S \cup \{s'\}
\]

The augmented test suite \( S' \) now includes sentences that exhibit the desired feature combinations, leading to a more comprehensive evaluation of the model's capabilities. Additionally, as stated in the evaluation report, these types of situations might occur in user experiences and are viewed as real-life scenarios \cite{misleading}. Therefore, it is crucial to include these generated tests in the test set that showcases these real-life, uncovered scenarios.

For the purpose of generating test cases, we selected the Claude language model \cite{anthropic2023} due to its capacity to generate varied and innovative test sets that are in line with our research goals. This model's attributes enable the generation of test cases that meet research requirements and highlight the robustness and reliability of sentiment analysis.

The generated sentences lack an emotion label, making them unsuitable for use as a test. In order to assign labels to the generated tests, we utilized six of the most accurate sentiment analysis models that were identified in the research literature \cite{savani2023bertbaseuncasedemotion,savani2023bertweetbasefinetunedemotion,savani2023robertabaseemotion,bergum2023xtremedistilemotion,jorgeutd2023sagemakerrobertabaseemotion,emanuel2023bertweetemotionbase}. with their accuracy mentioned in table~\ref{t3}. The input sentences were fed into these models for labeling, and the resulting labels were multiplied by their accuracy on the dataset (coefficient). The sentiment with the highest score was assigned as the label for each generated sentence, a method known as ``Differential Testing''~\cite{diff}. The system under test was then evaluated using a combination of original and newly generated tests to assess its accuracy on the updated test set.
\begin{table}[t]
\centering
\caption{Model performances under the same dataset}
\label{t3}
\scriptsize
\begin{tabular}{|l|c|c|c|}
\hline
Model & Accuracy & Precision & Loss \\
\hline
roberta-base-emotion & 0.931 & 0.936 & 0.152 \\
bert-base-uncased-emotion & 0.926 & 0.927 & 0.173 \\
bertweet-base-finetuned-emotion & 0.929 & 0.928 & 0.182 \\
xtremedistil-emotion & 0.926 & 0.928 & 0.226 \\
bertweet-emotion-base & 0.928 & 0.929 & 0.135 \\
sagemaker-roberta-base-emotion & 0.931 & 0.934 & 0.174 \\
\hline
\end{tabular}
\end{table}
\section{Experimental Results and Discussion}\label{ssec:evaluation}

\begin{figure*}[!ht]
  \centering
  \begin{subfigure}[b]{0.495\textwidth}
    \centering
    \includegraphics[width=\textwidth]{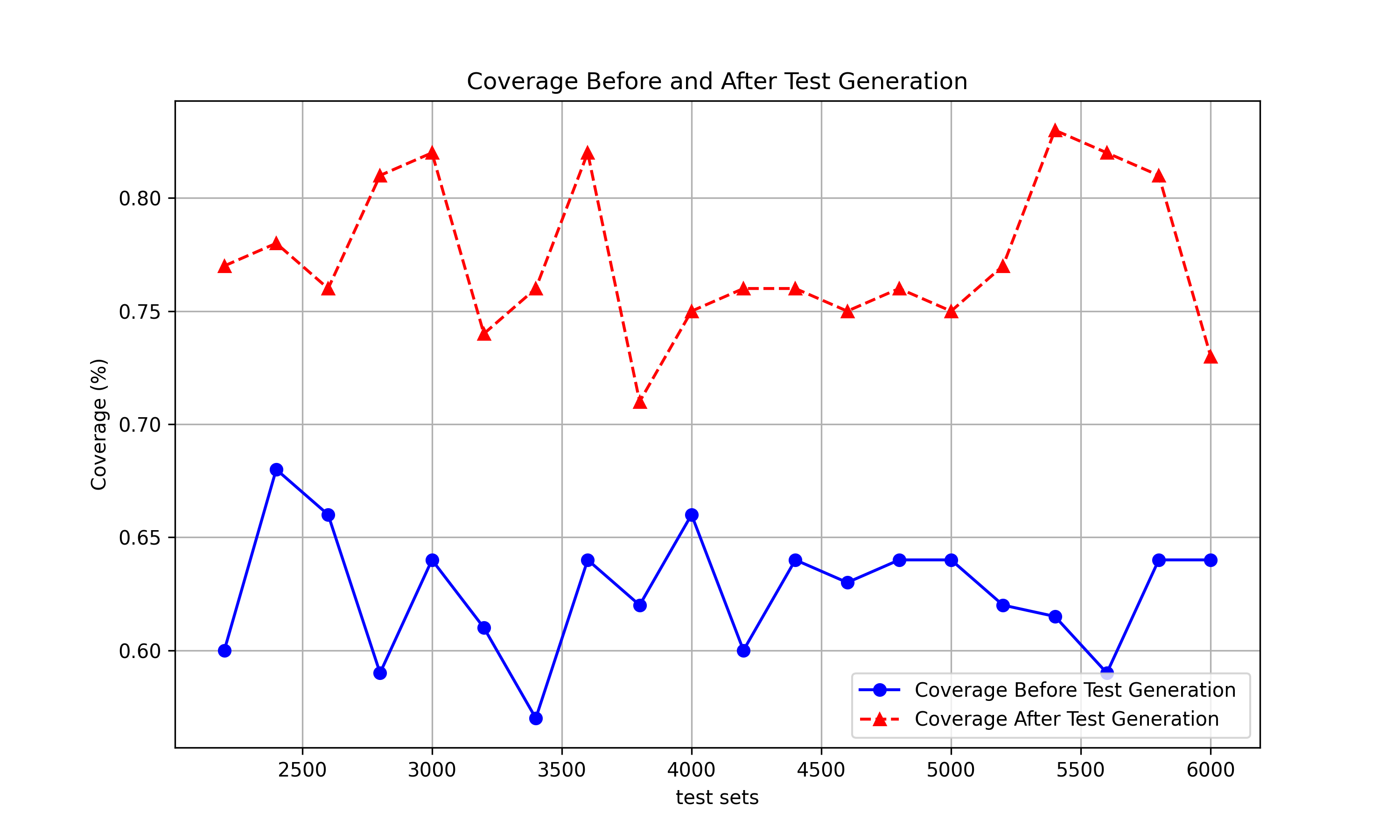}
    \caption{Test Case Generation: Impact on Coverage for $k$=2}
    \label{fig:mesh1}
  \end{subfigure}
  \hfill
  \begin{subfigure}[b]{0.495\textwidth}
    \centering
    \includegraphics[width=\textwidth]{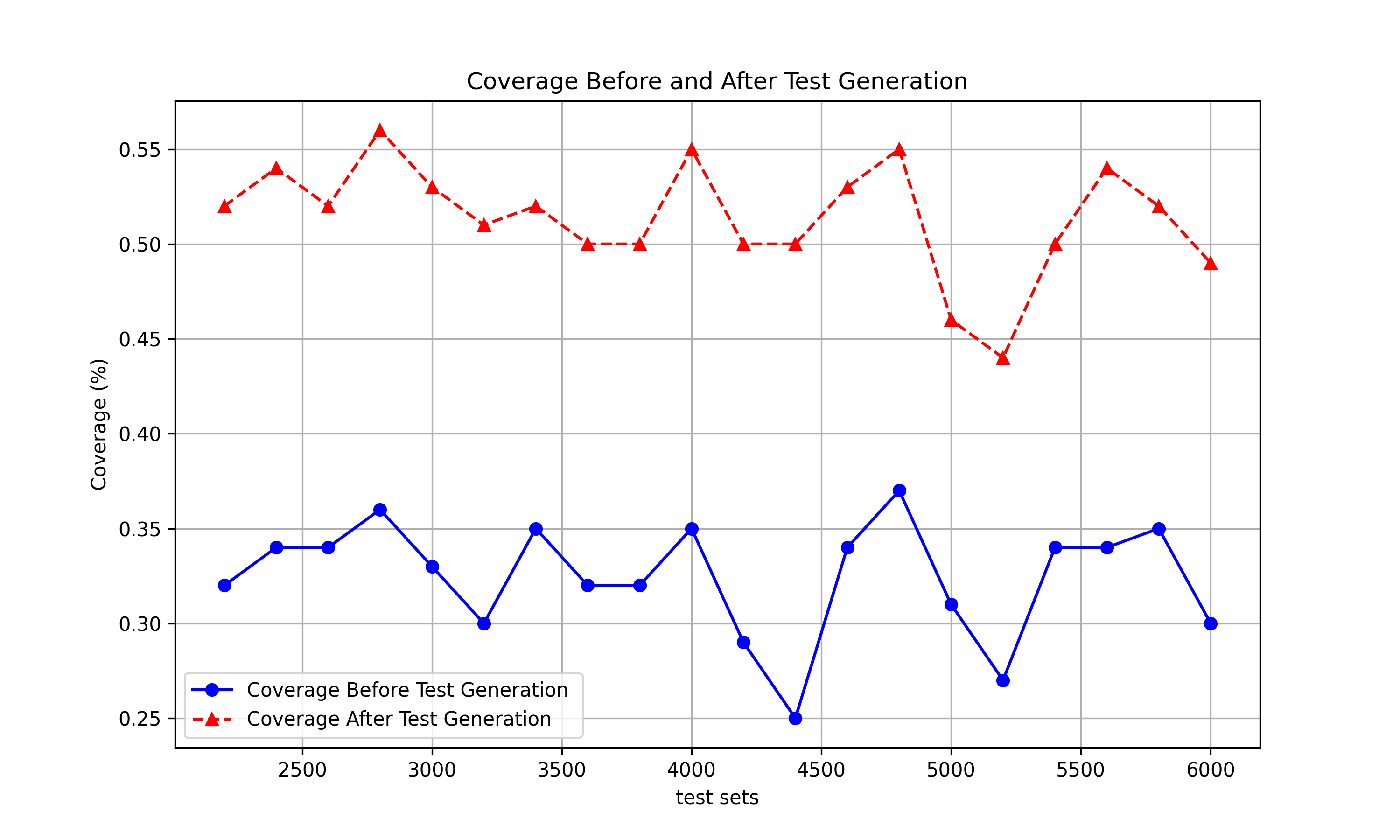}
    \caption{Test Case Generation: Impact on Coverage for $k$=3}
    \label{fig:mesh2}
  \end{subfigure}
  \caption{Coverage comparison for $k$=2 and $k$=3}
  \label{fig:sidebyside1}
\end{figure*}

\begin{figure*}[!ht]
  \centering
  \begin{subfigure}[b]{0.495\textwidth}
    \centering
    \includegraphics[width=\textwidth]{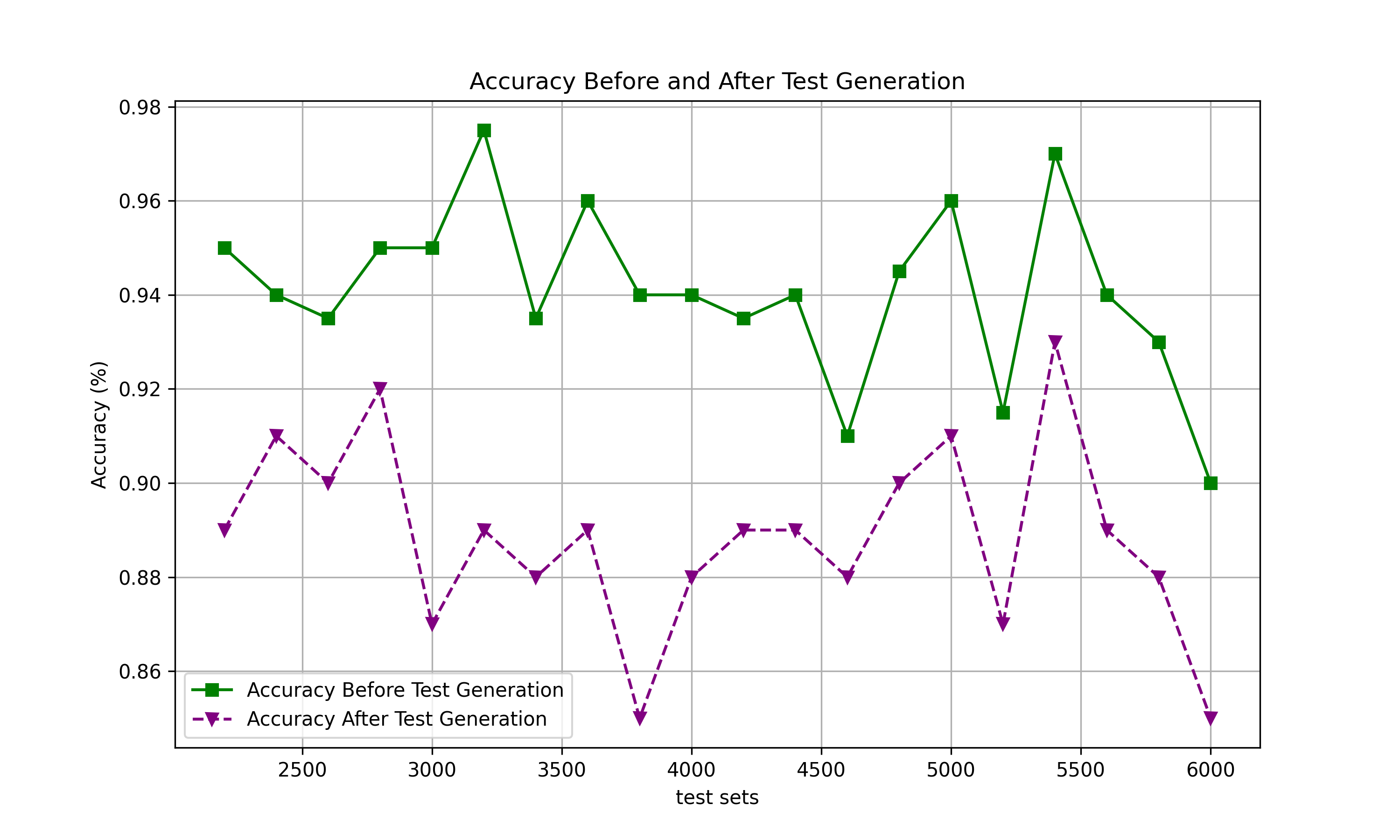}
    \caption{Test Case Generation: Impact on Accuracy for $k$=2}
    \label{fig:mesh3}
  \end{subfigure}
  \hfill
  \begin{subfigure}[b]{0.495\textwidth}
    \centering
    \includegraphics[width=\textwidth]{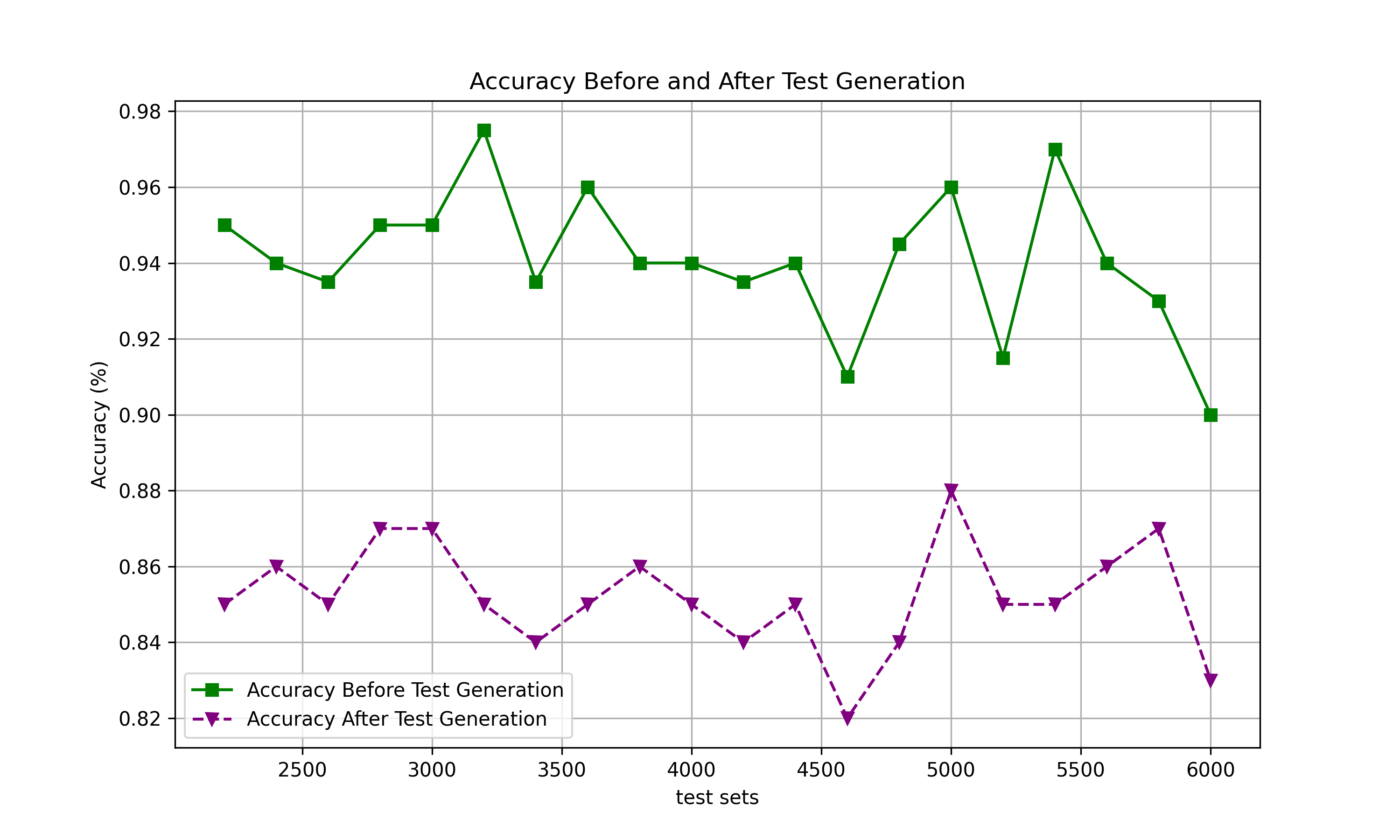}
    \caption{Test Case Generation: Impact on Accuracy for $k$=3}
    \label{fig:mesh4}
  \end{subfigure}
  \caption{Accuracy comparison for $k$=2 and $k$=3}
  \label{fig:sidebyside2}
\end{figure*}

The dataset utilized for assessing the results is denoted as \emph{CARER} \cite{Carer}. This dataset contains six primary emotions expressed through text, enabling a thorough assessment of the neural network's ability to accurately classify emotions.
The dataset is split into training and testing sets. The training set comprises 16,000 tweets, each associated with six output labels. Similarly, the test set consists of 2,000 tweets, each labeled with six output categories. A total of 18,000 tweets are chosen as the test set and assessed using the proposed coverage evaluation method. For this study, the dataset is divided into smaller subsets, each consisting of 200 tweets, in order to analyze the coverage capability of each subset. Additional tests will be set based on the uncovered features.

Our model is compared to the existing test dataset since, based on our findings, no other model conducts test generation for neural networks in text categorization in the same way. This study applied the Claude language model \cite{anthropic2023} for test generation and 6 sentiment analysis models for differential testing \cite{savani2023bertbaseuncasedemotion, savani2023bertweetbasefinetunedemotion, savani2023robertabaseemotion, bergum2023xtremedistilemotion, jorgeutd2023sagemakerrobertabaseemotion, emanuel2023bertweetemotionbase}. A distinct arrangement of the \emph{DISTILBERT} model \cite{bhadresh_savani_distilbert_base_uncased_emotion} was employed for word-level analysis.The process of learning and loading was executed on Google Colab, utilizing a central processing unit (CPU) with a capacity of 12.7 gigabytes of random access memory (RAM).

We tried different values of $k$ in our implementation to evaluate the effect of $k$ in our approach.
The results presented in Fig. \ref{fig:sidebyside1} illustrate the differences in coverage among various implementations of $k$. When $k$=2, the coverage increases on average by 14\% (Fig. \ref{fig:mesh1}). However, when $k$=3, the average increase in coverage is 18\% (Fig. \ref{fig:mesh2}).
The initial coverage for $k$=3 is 32\%, while the initial coverage for $k$=2 is 62\%. This indicates that the test suite lacks sufficient coverage for complicated sentences that involve three combinations of features. Although full coverage was not attained, the sentences generated noticeably raised the total test coverage in comparison to the original data. 

Figure \ref{fig:sidebyside2} demonstrates a decline in accuracy for both test generation implementations, suggesting that the generated tests expose previously unexplored areas and have the potential to improve the model's functionality.
When $k$=2 (Fig. \ref{fig:mesh3}), the accuracy experiences an average loss of 5\%, whereas for $k$=3 (Fig. \ref{fig:mesh4}), there is an average decrease of 8\%. These findings indicate that the tests created for $k$=3 demonstrate a higher number of defects in the model's decision-making capabilities.

The average accuracy after test generation for $k$=3 models is 85\%, whereas for $k$=2 models it is 88\%, which is lower than the model's accuracy on the base test suite, which is 94\%. This indicates a decrease in the model's precision when tested on these more challenging data points. This drop results from intentionally pushing the model to its limits in order to uncover areas for enhancement.
Based on the experimental results for  \( k = 3 \) the augmented test suite exhibits a greater decrease in accuracy compared to  \( k = 2 \). This can be attributed primarily to the production of more intricate sentence structures. Nevertheless, when \( k = 3 \), there is a greater time complexity and increased costs involved in generating these sentences. The sentences produced for \( k = 3 \) exhibit a significant level of complexity and deviate considerably from real-life situations, thereby making it challenging to evaluate the model's performance in practical settings.

\section{Conclusion and Future Works}\label{sec:conclusion}
This study introduced novel coverage criteria for evaluating test suites of transformer-based sentiment analysis neural networks. By using input space partitioning focused on emotionally relevant linguistic features and the k-projection coverage metric, we efficiently developed test cases and analyzed key feature subsets.
Our methodology demonstrated a 16\% average increase in test coverage for all methods and a 6.5\% decrease in model accuracy on augmented tests. 
The proposed criteria balance bug detection capability and computational efficiency, surpassing existing techniques. By emphasizing emotional dimensions through input space partitioning, our approach ensures thorough evaluation across diverse real-world scenarios.

Future work could extend our coverage criteria to other transformer-based NLP tasks like text summarization or machine translation. Expanding linguistic elements beyond verbs, adjectives, adverbs, and nouns could improve the detection of subtle emotional nuances, including negations and emojis. Finally, investigating the scalability and efficiency of our methodology on larger datasets and wider applications will be essential for real-world, industry-scale applications.
\bibliographystyle{IEEEtran}
\bibliography{main} 
\end{document}